\documentclass[sigconf]{acmart}
\pdfoutput=1
\usepackage{algorithm}
\usepackage{algpseudocode}
\usepackage{tikz}
\usepackage{pgfplots}
\usepackage{multirow} 
\usepackage{multicol} 
\usepackage{arydshln}
\usepackage{sidecap}
\usepackage[shortlabels]{enumitem}
\def\BibTeX{{\rm B\kern-.05em{\sc i\kern-.025em b}\kern-.08emT\kern-.1667em\lower.7ex\hbox{E}\kern-.125emX}}
    
\copyrightyear{2019} 
\acmYear{2019} 
\setcopyright{acmcopyright}
\acmConference[MMAsia '19]{ACM Multimedia Asia}{December 15--18, 2019}{Beijing, China}
\acmBooktitle{ACM Multimedia Asia (MMAsia '19), December 15--18, 2019, Beijing, China}
\acmPrice{15.00}
\acmDOI{10.1145/3338533.3366583}
\acmISBN{978-1-4503-6841-4/19/12}

\begin{document}

%
\title{Weakly Supervised Video Summarization  \\ by Hierarchical Reinforcement Learning}

\renewcommand{\shorttitle}{Weakly Supervised Video Summarization by Hierarchical Reinforcement Learning}

\author{Yiyan Chen, Li Tao, Xueting Wang, Toshihiko Yamasaki}
\affiliation{%
  \institution{The University of Tokyo}
  \city{Tokyo}
  \country{Japan}
}
\email{{chenyiyan, taoli, xt_wang, yamasaki}@hal.t.u-tokyo.ac.jp}

\begin{abstract}
Conventional video summarization approaches 
based on reinforcement learning have the problem 
that the reward can only be received 
after the whole summary is generated.
Such kind of reward is sparse 
and it makes reinforcement learning hard to converge. 
Another problem is that labelling each frame 
is tedious and costly, 
which usually prohibits the construction of 
large-scale datasets. 
To solve these problems, we propose a weakly supervised hierarchical reinforcement learning framework, 
which decomposes the whole task 
into several subtasks to enhance the summarization quality. 
This framework consists of a manager network 
and a worker network. 
For each subtask, the manager is trained to set a subgoal 
only by a task-level binary label, which requires much fewer labels than conventional approaches. 
With the guide of the subgoal, 
the worker predicts the importance scores for video frames 
in the subtask by policy gradient according to 
both global reward and innovative defined sub-rewards 
to overcome the sparse problem. 
Experiments on two benchmark datasets 
show that our proposal has achieved the best performance, even 
better than supervised approaches.
\end{abstract}

%
%



%

%

%
\maketitle
\begin{figure}[h]
  \centering
  \includegraphics[width=0.9\linewidth]{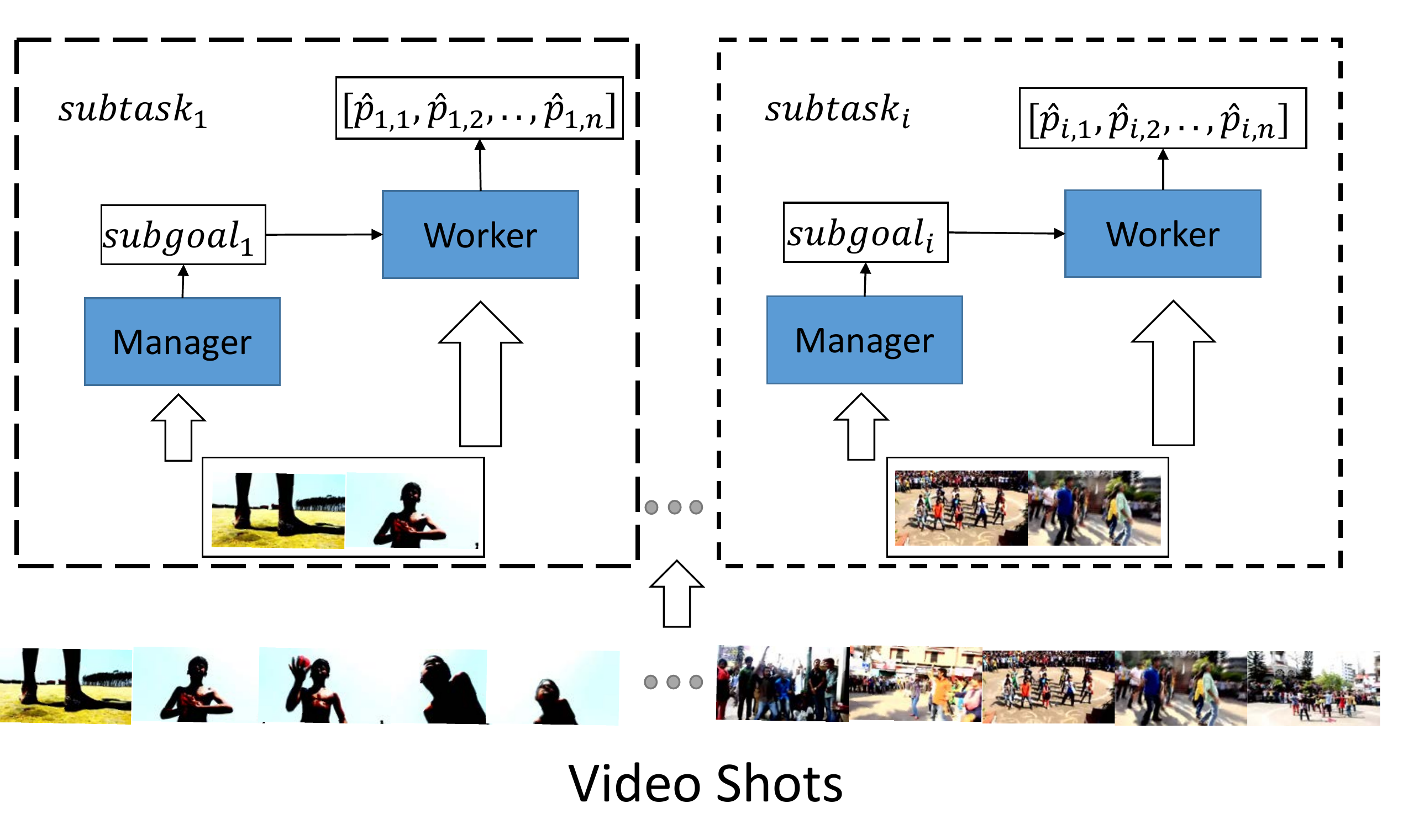}

  \caption{The concept of the proposed hierarchical 
  reinforcement learning based video summarization. 
  The total summarization task is divided into several subtasks, where each 
  subtask processes on several video frames. 
  A Manager is designed to set a subgoal 
  for each subtask, and a Worker is guided by it to predict the importance score for each frame.}
  \Description{task to subtask}
 
  \label{mainIdea}
\end{figure}

\section{Introduction}
Nowadays, video data are increasing explosively on the Internet. 
Video processing has attracted much attention from researchers. 
However, considering the temporal information 
for a long duration is still extremely challenging. 
A video can be considered as a sequence of frames.
Usually, we can downsample the sequence of frames. 
But it is still too long and it is a big problem for most video tasks.  
Video summarization, which shortens an original video to a compact summary,can offer us a short and representative sequence of features~\cite{kts}\cite{vs_0}. 

Recently, deep neural networks have been applied to video summarization. 
Recurrent neural networks are used to extract the temporal information~\cite{vs_lstm}\cite{hsa_rnn}. 
These supervised methods require a large number of frame-level labels for each video. 
Consequently, collecting a large scale of annotated videos costs a lot. 
The annotations for small temporal interval are also prone to be subjective and diverse. 
Moreover, some researchers propose unsupervised methods such as~\cite{dr_dsn}. 
They propose an unsupervised reinforcement learning method 
and train the model by policy gradient with a diversity-representativeness global reward to evaluate the selected frames. 
However, the reward can only be obtained after the whole summary is generated.
This kind of global reward is too sparse to evaluate a long series of actions. 
Here, each action is defined as whether to select a frame or not. 

Therefore, we propose a new weakly supervised video summarization method featuring hierarchical reinforcement learning, 
which requires only a small number of annotations and avoids the sparse reward problem. 
The main idea is to divide the whole task into several subtasks, 
while each subtask includes several sub sequences of frames divided from the whole sequence in order. 
In this way, only task-level annotations (detail in Sec. 3.3), i.e., 
weakly annotated labels which is given to each subtask instead of annotating all the frames, are required. 
The sparse reward problem can also be solved by setting a subgoal for each subtask. 
As shown in Figure~\ref{mainIdea}, our model consists of a Manager, 
which sets the subgoal, and a Worker, 
which predicts the importance score for each frame of the subtask in order to achieve the subgoal. 
Besides the global reward proposed in~\cite{dr_dsn}, we define a sub-reward for each subtask to train the Worker. 
The summary can then be generated by selecting frames according to their importance scores with a given limit of the video length. 
Experiments on two benchmark datasets show improvement compared to the state-of-art approaches~\cite{sum_gan}\cite{dr_dsn}. 
The margin from the previous works is very significant 
when we employ the new evaluation measure for video summarization 
that has been proposed in~\cite{Otani2019RethinkingTE}. 
Overall, the main contributions of this paper are as follows: 

\begin{itemize}[nosep]
  \item To the best of our knowledge, 
  we are the first to apply hierarchical reinforcement learning 
  to video summarization. 
  We solve the sparse reward problem by defining subgoals and sub-rewards.
  \item Our proposal is weakly supervised and requires only task-level annotations.  
  In this way, the cost on annotation and the subjective
  difference can be reduced.
  \item Our proposal not only outperforms state-of-the-art unsupervised methods 
  but also the supervised extensions of them.
  \end{itemize}

\section{Related Work}
\subsection{Video Summarization}
Video summarization is an important topic in video processing. 
Recently, researches on video summarization have been explored extensively 
and achieved great advances. These methods can be roughly divided 
into supervised methods, unsupervised methods, and weakly supervised methods.

Supervised methods use the frame-level annotations to 
train a model to predict the importance scores. 
Recurrent neural networks have been applied to capture the temporal information. 
\cite{vs_lstm} used long short-term memory (LSTM) to predict the importance score 
with a Determinantal Point Process (DPP) module.
\cite{hsa_rnn} proposed a hierarchical structure 
which took frames as input 
and considered the segmentation process first.
Performance of supervised methods can be improved by training with more annotated data. 
However, the cost is high to collect numerous videos with frame-level annotations.
 
Unsupervised methods with well-designed criteria 
require no annotation. 
\cite{sum_gan} proposed an unsupervised method by using generative adversarial networks (GANs). 
The discriminator is designed to compare the generated summary and the original video. 
\cite{dr_dsn} applied reinforcement learning to unsupervised video summarization. 
Instead of utilizing the annotations by humans, 
they defined a global reward considering the diversity and representativeness of the summary, 
and thus, the generated summary could be less subjective. 

Weakly supervised methods required a small number of annotations 
and could achieve great performance.   
\cite{Cai2018WeaklySupervisedVS} proposed a weakly supervised method 
that only required the topic label for a video.
A variational autoencoder (VAE) model was trained by the massive edited videos with topic labels 
on the Internet to learn a better video representation.
Their summarization network was implemented by a unified conditional variational encoder-decoder. 

\subsection{Reinforcement Learning}
Reinforcement learning has been explored a lot in recent years. 
It focuses on how the agent interacts with an environment. 
The agent is trained by a reward given by the environment. 
\cite{atari_drl} used a deep neural network 
to approximate the Q function 
and achieved great performance on many Atari games. 
Then, deep reinforcement learning has been applied into many different fields 
including computer vision~\cite{Huang2017LearningPF,Yu2018CraftingAT,aaai_furuta_2019}. 

If the reward can only be received after a series of actions, 
the sparse reward problem could make it difficult and inefficient to train a model~\cite{intrin_hrl}. 
Practically, this problem exists when applying reinforcement learning in many fields, 
such as text generation, natural language processing~\cite{drl_nlp}\cite{info_rl}
and object tracking~\cite{zhang2017deep}\cite{Ren_2018_ECCV}. 

Hierarchical reinforcement learning is a promising technology to solve this problem. 
The basic idea is to define the agent as multi-layer structures dealing with 
different subtasks to learn several sub policies. 
The macro policy is learned to choose different sub policies for subtasks. 
This kind of method can achieve a great performance effectively 
with the help of domain knowledge to divide the task and define the subtasks well. 
\cite{fn_hrl} proposed a flexible and end-to-end framework 
for hierarchical reinforcement learning. They divided an agent into a Manager and a Worker. 
The Manager sets abstract goals which are conveyed to and enacted by the Worker. 
And the Worker takes actions to interact with the environment. 

\begin{figure*}[h]
  \centering
  \includegraphics[width=1.0\linewidth]{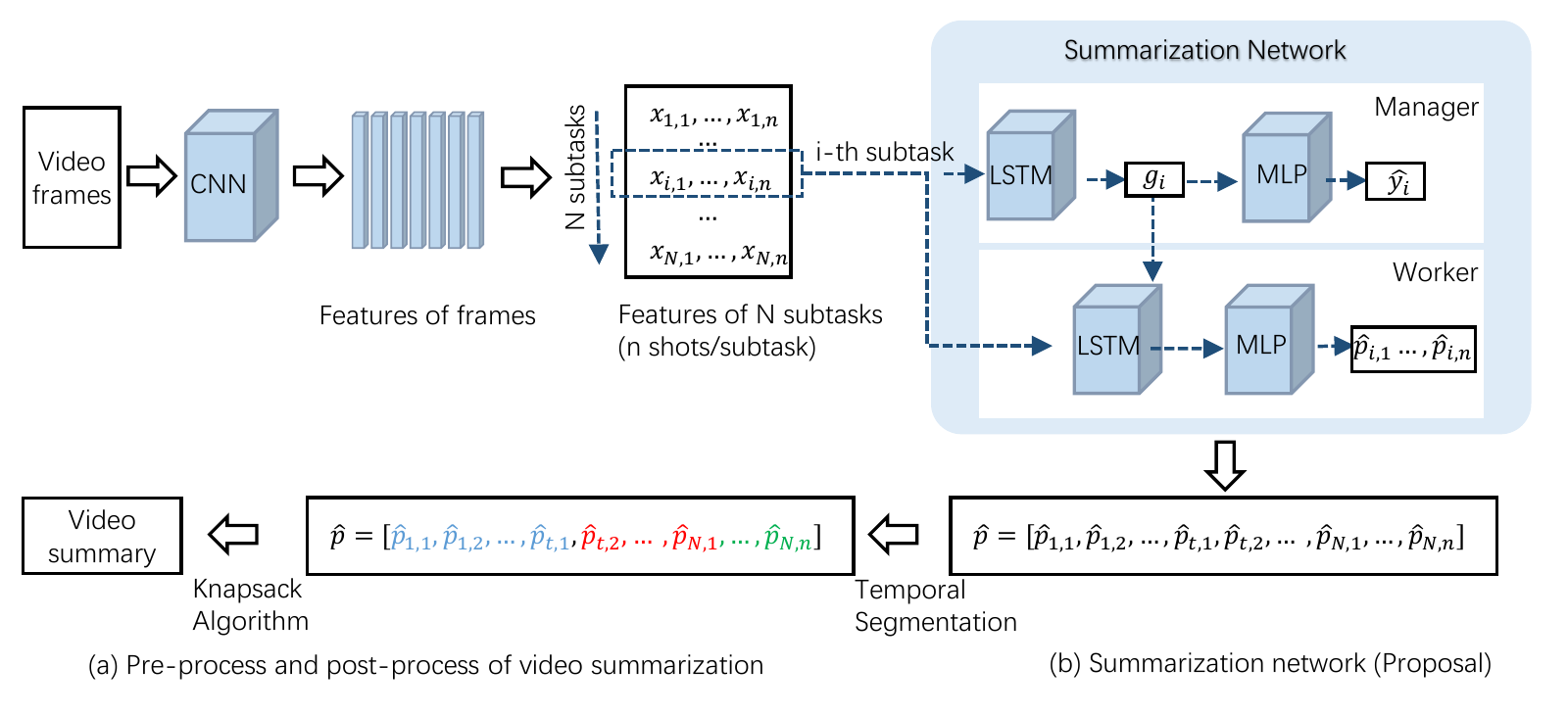}

  \caption{(a) is the general pre-process and post-process for video summarization. 
  (b) is the proposed hierarchical reinforcement learning model including the Manager and the Worker.  
  }

  \label{vs framework}
\end{figure*}

\section{Proposed Method}
Our proposed method regards the whole video summarization task as the combination of subtasks.  
The hierarchical architecture consists of a Manager and a Worker.
The Manager sets a subgoal for each subtask.
The Worker takes its action following the subgoal.
In the following discussion, we define that the whole task is separated into $N$ subtasks. 
Each subtask processes $n$ frames. 
The whole task includes $N \times n$ frames. 
For each step, the input of the Manager is a sequence of frames in one subtask with size $n$, represented as $[x_{i,1}, x_{i,2},...,x_{i,n}]$.
The Manager sets a subgoal $g_i$ ($i\in[1,N]$). 
Conditioned on the subgoal $g_i$, the Worker predicts the importance score $\hat{p}_{i,t}$ ($t \in [1,n]$) for each frame within $i$-th subtask. 
The whole task can be achieved better when the subtasks are achieved well.
Our work follows the most common procedures of video summarization task.
Therefore, we will introduce the basic procedures first and then focus on the proposal.

\subsection{Basic Procedures}
Video summarization converts an original long video 
into a short and concise summary. 
As shown in Figure~\ref{vs framework}, the whole framework consists of several procedures including pre-processing, 
summarization network, and post-processing. 

A pre-trained CNN network is used to extract the frame features. \cite{vs_lstm} provided a dataset that uses GoogLeNet to extract the frame features. 
For a fair comparing, we follow them to obtain the sequence of features and subsample by 2fps.

The summarization network is used to predict the importance score for each frame. 
In order to promise the generated summary to be natural and less flashed, temporal segmentation is usually used to divde the sequence of frames into different groups as shots. 
The most common temporal segmentation method is a kernel temporal
segmentation (KTS) algorithm, which partitions the sequence of
frames into shots considering the similarity among frame features~\cite{kts}. 
The importance score of each shot is calculated as the average score of the frames
included in the shot. We select the shot with higher importance score under a given limit of the video length. 
Considering that each shot includes a variable number of frames, this selection is of the knapsack problem 
and we follow~\cite{dr_dsn} using a near-optimal solution by dynamic programming~\cite{vs_0} as the post-processing.

\subsection{Hierarchical Structure}
As shown in Figure~\ref{vs framework}(b), the proposed summarization network uses reinforcement learning with a hierarchical architecture. 
Note that the hierarchical reinforcement learning is already proposed in~\cite{fn_hrl}, 
but this is the first work to apply hierarchical reinforcement learning to video summarization to the best of our knowledge. 
Besides, how to divide the whole task into a hierarchical manner requires careful design, 
which is also our technical contribution. 
Our proposal pays attention not only to the quality of the whole generated summary, 
but also to the quality of subtasks.
For the quality of the whole summary, we apply a widely used global reward defined as diversity-representativeness reward 
proposed by~\cite{dr_dsn}.
Then, we regard the total summarization task as several subtasks and 
design an agent with the hierarchical structure using a kind of sub-reward representing the quality of the subtask. 
The agent consists of two recurrent neural networks: one is called Manager and the other is called Worker. 
The Manager sets the subgoal to each subtask. 
In order to achieve the corresponding subgoal, the Worker deals with the frames within the subtask. 
Both Manager and Worker are implemented by LSTM. 
We train the Manager with a smaller number of ground-truth annotations compared to supervised methods, 
and the Worker with the REINFORCE algorithm~\cite{Williams1992}, which belongs to a family of reinforcement learning methods.

Then, we introduce the Manager network, the Worker network, and two kinds of rewards in detail as follows.

\subsection{Manager Network}
To train the Manager, we define the task-level label $y_i$ as 
$1$ if there exists one key frame in a sub sequenceis and $0$ otherwise. 
\begin{equation}   
  y_i = \begin{cases}
    1 & \text{if } p_{i,t} =1, \exists{t} \in \left[ 1,n \right] \\
    0 & \text{otherwise}
  \end{cases},
   \label{y_i}
  \end{equation}
where $p_{i,t}$ is the ground-truth label for each frame. 

The Manager consists of an LSTM and multilayer perceptron (MLP). 
As mentioned above, 
for each step the Manager takes the sub sequence of  $[x_{i,1},x_{i,2},...,x_{i,n}]$ ($n$ is set empirically) 
as input. And then we take the last hidden state as the subgoal $g_i$ of the $i$-th subtask. 
\begin{equation}
  h_{i,t} = f_{\beta_m}\left( x_{i,t},h_{i,t-1} \right) , x_{i,t} \in \left[ x_{i,1}, x_{i,2}, ..., x_{i,n} \right], 
\end{equation}
\begin{equation}
  g_i = h_{i,n},  
\end{equation}
\begin{equation}
  \hat{y_i} = \text{sigmoid} \left( w_m \cdot g_i + b_m \right),  
  \label{haty_i}
\end{equation}
where $f_{\beta_m}$ represents the LSTM of the Manager with parameter $\beta_m$, 
$w_m$ and $b_m$ are parameters of MLP, $\hat{y_i}$ is the predicted probability of whether the $i$-th sub sequence of frames includes a key frame and $h_{i,t}$ is the hidden state ($h_{i,0} = h_{i-1,n}$ and $h_{0,0}$ is zero-initialized). 

\subsection{Worker Network}
The Worker is also composed of an LSTM and MLP.  
It takes a sub sequence of frames as input. 
Then, it outputs the importance score for each frame conditioned on the subgoal 
given by the Manager. 
Our Worker takes a sub sequence of frames 
$\left[ x_{i,1},x_{i,2},…,x_{i,n} \right]$ ($n$ is set empirically) as input  
and a subgoal $g_i$ given by the Manager. 
The Worker predicts the importance scores for each frame.
\begin{equation}
  h_{i,t} = f_{\beta_w}\left( x_{i,t},h_{i,t-1} \right), x_{i,t} \in \left[ x_{i,1},x_{i,2},...,x_{i,n} \right], 
\end{equation}
\begin{equation}
  h^{'}_{i,t} = w_{1}\left[ g_i,h_{i,t} \right] + b_{1}, t \in \left[ 1,n \right], 
\end{equation}
\begin{equation}
  \hat{p}_{i,t} = \text{sigmoid} \left( w_w \cdot h^{'}_{i,t} + b_w \right), 
\end{equation}
where $f_{\beta_w}$ represents the LSTM with parameter $\beta_w$, 
$h_{i,t}$ is the hidden state ($h_{i,0} = h_{i-1,n}$ and $h_{0,0}$ is zero-initialized), $w_{1}$ and $b_{1}$ are parameters of a linear layer 
that combines the subgoal and the hidden state,  
$w_w$ and $b_w$ are parameters of MLP, $\left[ g_i,h_{i,t} \right]$ is to concatenate two vectors 
and $\hat{p}_{i,t}$ is the predicted importance score for frame $x_{i,t}$. 

After predicting the importance score of each frame, 
we regard the importance score as the probability parameter of a Bernoulli distribution.
This determines the probability of a frame being selected for the summary.
\begin{equation}a_{i,t} \sim \text{Bernoulli}\left( \hat{p}_{i,t} \right),a_{i,t}=0\text{ or } 1, \end{equation}
where $a_{i,t}=1$ is the action and means that $x_{i,t}$ is selected in the summary.

\subsection{Reward Function}
\cite{dr_dsn} defines the diversity-representativeness reward $R_{dr}$. 
This reward includes diversity reward $R_d$ and representativeness reward $R_{rep}$.
$R_d$ is computed by summing up the dissimilarity of frame feature within the selected subset. 
$R_{rep}$ is obtained by considering 
the minimal distance 
between each frame feature with others within the selected subset.
\begin{equation}
  R_d = \frac{1}{|Y|(|Y|-1)} \sum_{t\in Y} \sum_{ {t'} \in Y,{t'} \neq t}d\left( x_t,x_{t'} \right), 
\end{equation}

\begin{equation}
  d\left( x_t, x_{t'} \right) = 1 - \frac{{x_t}^T x{_t'}}{||x_t||_2 ||x_{t'}||_2},
\end{equation}

\begin{equation}
  R_{rep} = \text{exp}\left(  -\frac{1}{n \times N}\sum^{n\times N}_{t=1} \text{min}_{t'\in Y}||x_t - x_{t'}||_2 \right), 
\end{equation}

\begin{equation}
  R_{dr} = \frac{1}{2} R_d + \frac{1}{2}R_{rep}, 
\end{equation}
where $d\left( \cdot, \cdot\right)$ is the dissimilarity function and $t\in Y$ means that frame $x_t$ is selected into the summary.

This diversity-representativeness reward is given for a generated summary. 
Besides this, we also define a sub-reward $R_{sub}$ considering 
how our Worker achieves the subgoal. 
We compute the average of the outputs of our Worker 
and compare it with the importance score of the subtask predicted by our Manager. 
The sub-reward $R_{sub}$ is defined as follows:
\begin{equation}
  \hat{p_i} = \frac{1}{n} \sum^{n}_{t=1} \hat{p}_{i,t}, 
\end{equation}

\begin{equation}
  R_{sub} = \text{exp}\left( -\frac{1}{N} \sum^{N}_{i=1}||\hat{p_i} - \hat{y}_i||_2 \right), 
\end{equation}
where $\hat{p}_i$ is the average of the probabilities 
within a subtask, 
and we use an exponential function to rescale to get $R_{sub}$ as the sub-reward.

Therefore, the reward for the Worker is: $R=\alpha R_{dr}+\left( 1-\alpha \right) R_{sub}$, where 
$\alpha$ is the hyperparameter.

\begin{figure*}[t]

  \begin{center}
  \includegraphics[width=0.75\linewidth]{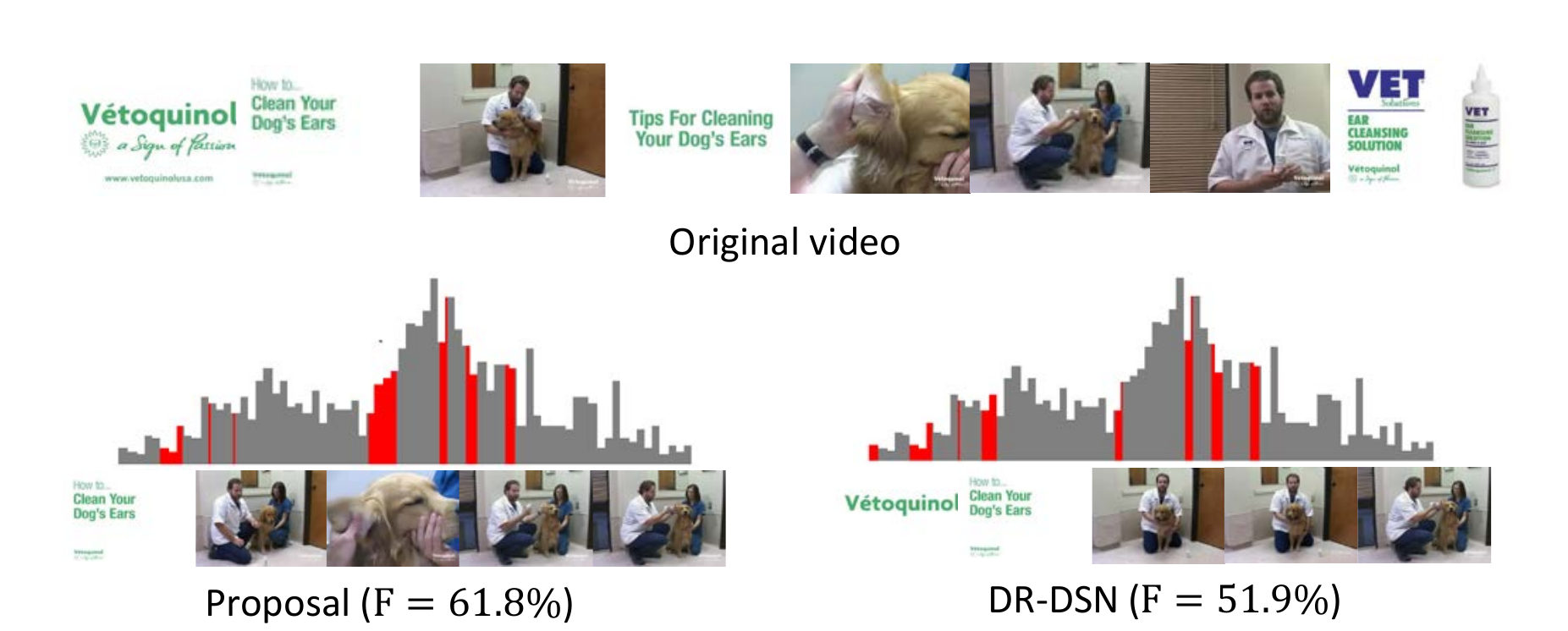}
  \end{center}
  
  \caption{Examples of summary videos of proposed method comparing with the state-of-the-art.}

  \label{selectedFrame}
\end{figure*}
\subsection{Optimization}
The loss function of the Manager is defined as follows:
\begin{equation}
L_m=-\frac{1}{N} \sum^N_{i=1} \  y_i \text{log}\left( \hat{y}_i \right)+\left( 1-y_i\right)\text{log}\left( 1-\hat{y}_i \right), 
\end{equation}
where $N$ is the number of subtasks 
which a video is divided into, and
$y_i$ and $\hat{y}_i$ are defined in Eq.~\ref{y_i} and Eq.~\ref{haty_i}, respectively.

The Worker is trained with REINFORCE algorithm~\cite{Williams1992}. 
It is aimed to learn a policy $\pi_{\theta_W}$ ($\theta_W$ is parameter of Worker in our case) 
through maximizing the expected rewards. 
\begin{equation}
J\left( \theta_W \right) = E_{p_{\theta_W}\left( a_{1:n \times N}\right) }\left[ R \right],
\end{equation}
where $p_{\theta_W}\left( a_{1:n \times N}\right) $ is the sequence of actions for the whole task.
Following the REINFORCE algorithm, we can compute 
the derivative of the objective function $J\left( \theta_W \right)$ in terms of $\theta_W$ as follows:
\begin{equation}
  \nabla_{\theta_W} J\left( \theta_W \right) = E_{p_{\theta_W}\left( a_{1:n\times N}\right)}\left[ R\sum^{n\times N}_{t=1} \nabla_{\theta_W} \text{log} \pi_{\theta_W} \left( x_t \right) \right].
\end{equation}
We approximate $R$ by letting the agent take action $c$ times.  
To avoid the high variance, we subtract a constant $b$ from the reward. 
\begin{equation}
  \nabla_{\theta_W} J\left( \theta_W \right)  \approx \frac{1}{c} \sum^c_{i=1}\left[ \sum^{n\times N}_{t=1} \left( R_i-b \right) \nabla_{\theta_W} \text{log} \pi_{\theta_W} \left(x_t\right)\right].
\end{equation}

\section{Experiments}
\subsection{Dataset}
Our proposal is evaluated on two benchmark datasets: 
SumMe~\cite{vs_1} and TVSum~\cite{tvsum}. 
SumMe includes 25 videos ranging from 1 to 6 minutes 
and for each video there are annotations from 15 to 18 users. 
TVSum includes 50 videos ranging from 2 to 10 minutes, 
and for each video there are annotations from 20 users. 
Both datasets provide frame-level importance scores. 

Besides, we also consider OVP\footnote{Open video project: https://open-video.org/} 
and YouTube~\cite{de2011vsumm} 
to evaluate our model following the settings by~\cite{vs_lstm} described in Section 4.3.

\subsection{Evaluation Metrics}
\textbf{F score}:
In order to compare the generated summary and the ground-truth, 
the most common evaluation approach~\cite{vs_0}\cite{vs_lstm}
is F score. 
Considering the intersection of two videos, 
the precision and recall are defined as follows:
\begin{equation}
Precision=\frac{A \bigcap B}{A},\text{ }Recall= \frac{A \bigcap B}{B}, 
\end{equation}
\begin{equation}
F=\frac{2\times Precision \times Recall}{Precision+Recall}, 
\end{equation}
where $A$ is the ground-truth summary and $B$ is the generated summary.

\noindent \textbf{Rank correlation coefficient}:
\cite{Otani2019RethinkingTE} proposed 
another evaluation metric considering rank order statistics.
For summarization methods based on importance scores, they claimed 
that the random methods can achieve comparable performance 
because of well-designed pre-processing and post-processing. 
Therefore, they reported that 
considering the importance score rank order 
between the prediction and the human annotation 
is a much better metric without the effect of post-processing.
They use Kendall's~$\tau$~\cite{10.2307/2332303} 
and Spearman's~$\rho$~\cite{citeulike:11910785} correlation coefficients.
\begin{figure}
  \centering
  \includegraphics[width=0.8\linewidth]{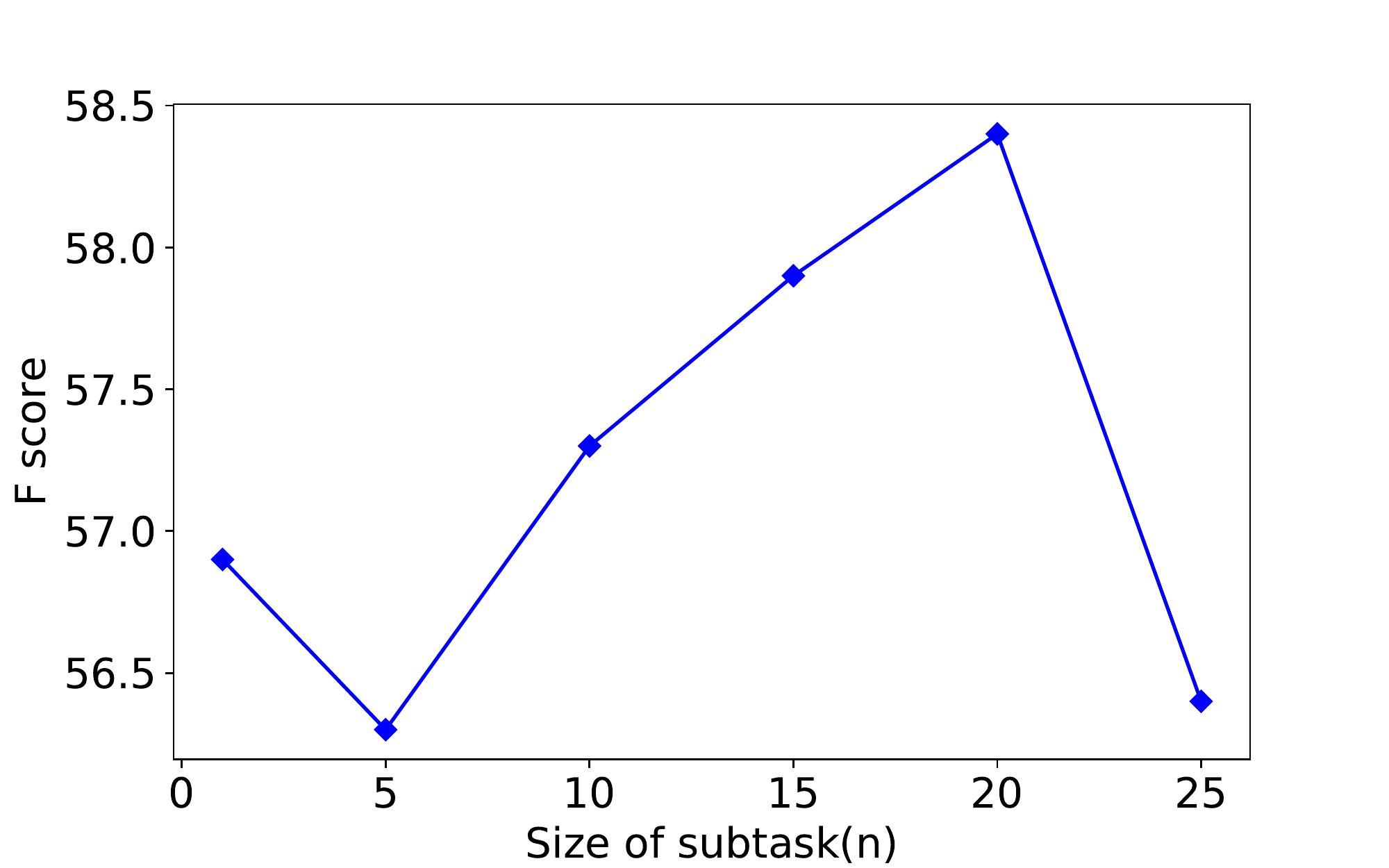}

  \caption{Experiments are conducted with different sizes of subtask on TVSum dataset.
  When $n=20$, it shows the best performance.
  }

  \label{FwithSizeTask}
\end{figure}

\subsection{Training}
There are three settings~\cite{vs_lstm}. (1) Canonical: 
we train our model by $5$-fold cross validation (5FCV). 
(2) Augmented: we use 5FCV as (1) with more training data of OVP and YouTube. 
(3) Transfer: for our target dataset (SumMe or TVSum), 
the other three datasets are used as training data.

We train the Manager and the Worker iteratively. We approximate the expected rewards
with $c = 10$ and set $\alpha = 0.5$. The size of a subtask is set as $n=20$. 
\subsection{Evaluation}
\begin{table*}[tp]  
  \centering
  \caption{Evaluation by F score for video summarization}

  \label{fresult}
  \begin{tabular}{|c|c|c|c|c|c|c|c|c|c|}
  \hline
  \multirow{2}*{} & \multirow{2}*{Methods} & \multicolumn{4}{c|}{SumMe} &  \multicolumn{4}{c|}{TVSum} \\ 
  \cline{3-10}
        &  & canonical & augmented & transfer & labels& canonical & augmented & transfer & labels\\
  \hline 
  \multirow{4}*{supervised} &vsLSTM~\cite{vs_lstm} & 37.6 & 41.6 & 40.7 & 293& 54.2 & 57.9 & 56.9 & 470 \\
  &dppLSTM~\cite{vs_lstm}  & 38.6 & 42.9 & 41.8 & 293&54.7 & 59.6 & 58.7 &470\\
  &SUM-GAN$_{\text{sup}}$~\cite{sum_gan} & 41.7&43.6 & -&293 &56.3 &{\bf 61.2} &- &470 \\
  &DR-DSN$_{\text{sup}}$~\cite{dr_dsn} & 42.1& 43.9 &{\bf 42.6} & 293 &58.1 &59.8 &{\bf 58.9} &470 \\
  \cdashline{1-10}[0.8pt/2pt]
  \multirow{2}*{unsupervised}
  &SUM-GAN~\cite{sum_gan}  & 39.1 & 43.4 & - & -&51.7 & 59.5 & - &-\\
  \cdashline{2-10}[0.8pt/2pt]
  &DR-DSN~\cite{dr_dsn}& 41.4 & 42.8 & 42.4 & - & 57.6 &58.4  &  57.8 & -\\

  \cdashline{1-10}[0.8pt/2pt]

  weakly supervised&{\bf Our Proposal}  & {\bf 43.6} & {\bf 44.5} & {42.4} & 15& {\bf 58.4} & {58.5} & {58.3} & 24\\
  \hline
  \end{tabular}

\end{table*}

To show that our subtasks help to improve the performance, 
we compare our model with the different sizes of subtask. 
As shown in Figure~\ref{FwithSizeTask}, 
we conduct the experiments on TVSum dataset. 
Our model shows the best performance when the size of subtask is set as $n=20$. 
It can be seen in Figure~\ref{FwithSizeTask} that when the size of subtask is 1, the performance is better than that of size 5. 
It is reasonable because we define the sub-reward using the average of the outputs of the Worker. 
Therefore, if the size of subtask is not large enough, the bias of the sub-reward could be large. 

We compare our proposal with previous supervised and unsupervised methods. 
vsLSTM and dppLSTM~\cite{vs_lstm} are supervised methods.
SUM-GAN~\cite{sum_gan} and DR-DSN~\cite{dr_dsn} are unsupervised methods.
SUM-GAN is based on GANs.
DR-DSN is based on reinforcement learning 
and the most related to our work. 
\cite{Cai2018WeaklySupervisedVS} is a weakly supervised method. 
However, they use a large number of web videos with category labels 
to learn more accurate and informative video representations. 
For a fair comparison, we don't include it here.
Table~\ref{fresult} shows the F score performance  
and the average number of labels for a video, which is required by each method.
Our hierarchical reinforcement learning based model 
achieves improvement on two benchmark datasets and requires much smaller number of task-level annotations than supervised methods. 

\begin{table}[t]
  \caption{Kendall's $\tau$ and Spearman's $\rho$ 
  correlation coefficients computed between different importance scores
  and human annotated scores. 
  }

    

  \begin{tabular}{cccc}
    \toprule
    
    \multirow{2}*{} & \multicolumn{2}{c}{TVSum} \\ 
    Methods&  $\tau$ &  $\rho$\\
    \midrule
    dppLSTM~\cite{vs_lstm}  & 0.042  & 0.055  \\
    DR-DSN~\cite{dr_dsn}  &  0.020 & 0.026  \\
    Proposal &  {\bf 0.078} & {\bf 0.116}\\
    \midrule
    Human  & 0.177 & 0.204  \\
  \bottomrule
\end{tabular}

\label{rankEvaluation}
\end{table}

On the canonical setting, our proposal outperformed the baseline dppLSTM by $5\%$ and $3.7\%$ points on two benchmark datasets.
And compared with the state-of-the-art DR-DSN, 
our proposal improves by $2.2\%$ and $0.8\%$ points. 
Our hierarchical structure requires a much smaller number of annotations (about $1/20$ in both cases) 
to train the Manager. 
The whole task is divided into several subtasks.
Compared to frame-level annotations, 
our proposal only requires task-level annotations 
to train the Manager. Therefore it is weakly supervised.
Then, we compared the performance with the supervised version of SUM-GAN and DR-DSN. 
On the canonical setting, our proposal outperformed SUM-GAN$_{\text{sup}}$ by $1.9\%$ and $2.1\%$ points on the two benchmark datasets.
Compared with DR-DSN$_{\text{sup}}$, the performance of our proposed method was improved by $1.5\%$ and $0.3\%$ points.
Our result was improved by $0.9\%$ and $0.1\%$ points with augmented data respectively on SumMe and TVSum. 
In particular, our proposal outperformed all the other methods on SumMe with augmented data. 
However, our proposal was not capable of the transfer task because subgoals between different domains may vary a lot.

Then, we evaluate the performance using the rank order statistics 
proposed in~\cite{Otani2019RethinkingTE} introduced in Sec. 4.2, which is claimed to be a better evaluation metric without the effect of post-processing. Compared to supervised and unsupervised methods in Table~\ref{rankEvaluation}, our proposal achieved higher correlation coefficients on both Kendall's~$\tau$ and Spearman's~$\rho$. Compared to supervised methods~\cite{vs_lstm},unsupervised methods~\cite{dr_dsn} improves the diversity by reinforcement learning but achieves lower correlation coefficients. With help of task-level labels, our proposal can promise the diversity and achieve higher correlation coefficients in the meantime.

An example is shown in Figure~\ref{selectedFrame}. 
The original video is about cleaning dog's ears. 
Our result skipped irrelevant video frames at the beginning of the video 
and selected frames that show the detail of how to clean the dog's ears. 
Therefore, our proposal achieved a higher F score.

\section{Conclusion}
In this paper, we proposed a weakly supervised hierarchical reinforcement learning method 
to avoid the sparse reward problem and improve the quality of video summarization.
Our experiments on two benchmark datasets showed 
that our hierarchical structure achieved the best performance out of all other methods known to us. 
In particular, evaluation using rank order statistics that was recently proposed in~\cite{Otani2019RethinkingTE} 
clearly showed the superiority of our proposed method.
Also, our proposal only requires a smaller number of 
task-level annotations to train the Manager.
In the future work, we will apply the proposed method to larger dataset, define an efficient label-free mechanism for the Manager, embed multi-modal and consider multi-field information such as audio.

\begin{acks}
This work was partially financially supported by the Grants-in-Aid for Scientific Research Numbers JP19H05590 and JP19K20289.
\end{acks}
%
\bibliographystyle{ACM-Reference-Format}
\bibliography{myreference}

%
\end{document}